\documentclass[preprint,3p,times,review]{elsarticle}

\usepackage{amssymb}
\usepackage{latexsym}
\usepackage{url}
\usepackage{xcolor}
\usepackage{hyperref}
\usepackage{booktabs}
\usepackage{caption}
\usepackage{multirow}
\usepackage{float}
\usepackage{tikz}
\usepackage{amsmath}
\usepackage{array}
\usepackage{wrapfig}

\usepackage{array}
\usepackage{booktabs}
\usepackage{makecell}
\usepackage[figuresright]{rotating}
\usepackage{bm}
\usepackage{comment}

\journal{Engineering Applications of Artificial Intelligence}

\begin{document}

\begin{frontmatter}

\title{Partitioned Memory Storage Inspired Few-Shot Class-Incremental learning}

\author[label1]{Renye Zhang}
\author[label2]{Yimin Yin}
\author[label3,label4]{Jinghua Zhang} \cortext[cor1]{Corresponding author}\ead{zhangjingh@foxmail.com}

\affiliation[label1]{organization={School of Computer Science, Hunan First Normal University},
             city={Changsha},
             state={Hunan},
             country={China}}
\affiliation[label2]{organization={School of Mathematics and Statistics, Hunan First Normal University},
             city={Changsha},
             state={Hunan},
             country={China}}
\affiliation[label3]{organization={Center for Machine Vision and Signal Analysis, University of Oulu}, 
            city={Oulu},
            country={Finland}}
\affiliation[label4]{organization={College of Intelligence Science and Technology, National University of Defense Technology},
             city={Changsha},
             state={Hunan},
             country={China}}

\begin{abstract}
Current mainstream deep learning techniques exhibit an over-reliance on extensive training data and a lack of adaptability to the dynamic world, marking a considerable disparity from human intelligence. To bridge this gap, \textit{Few-Shot Class-Incremental Learning} (FSCIL) has emerged, focusing on continuous learning of new categories with limited samples without forgetting old knowledge. Existing FSCIL studies typically use a single model to learn knowledge across all sessions, inevitably leading to the stability-plasticity dilemma. Unlike machines, humans store varied knowledge in different cerebral cortices. Inspired by this characteristic, our paper aims to develop a method that learns independent models for each session. It can inherently prevent catastrophic forgetting. During the testing stage, our method integrates \textit{Uncertainty Quantification} (UQ) for model deployment. Our method provides a fresh viewpoint for FSCIL and demonstrates the state-of-the-art performance on CIFAR-100 and \textit{mini}-ImageNet datasets.

\end{abstract}


\end{frontmatter}

\section{Introduction}\label{sec1}
\begin{figure}[h]
    \centering
    \includegraphics[width = 0.8\textwidth]{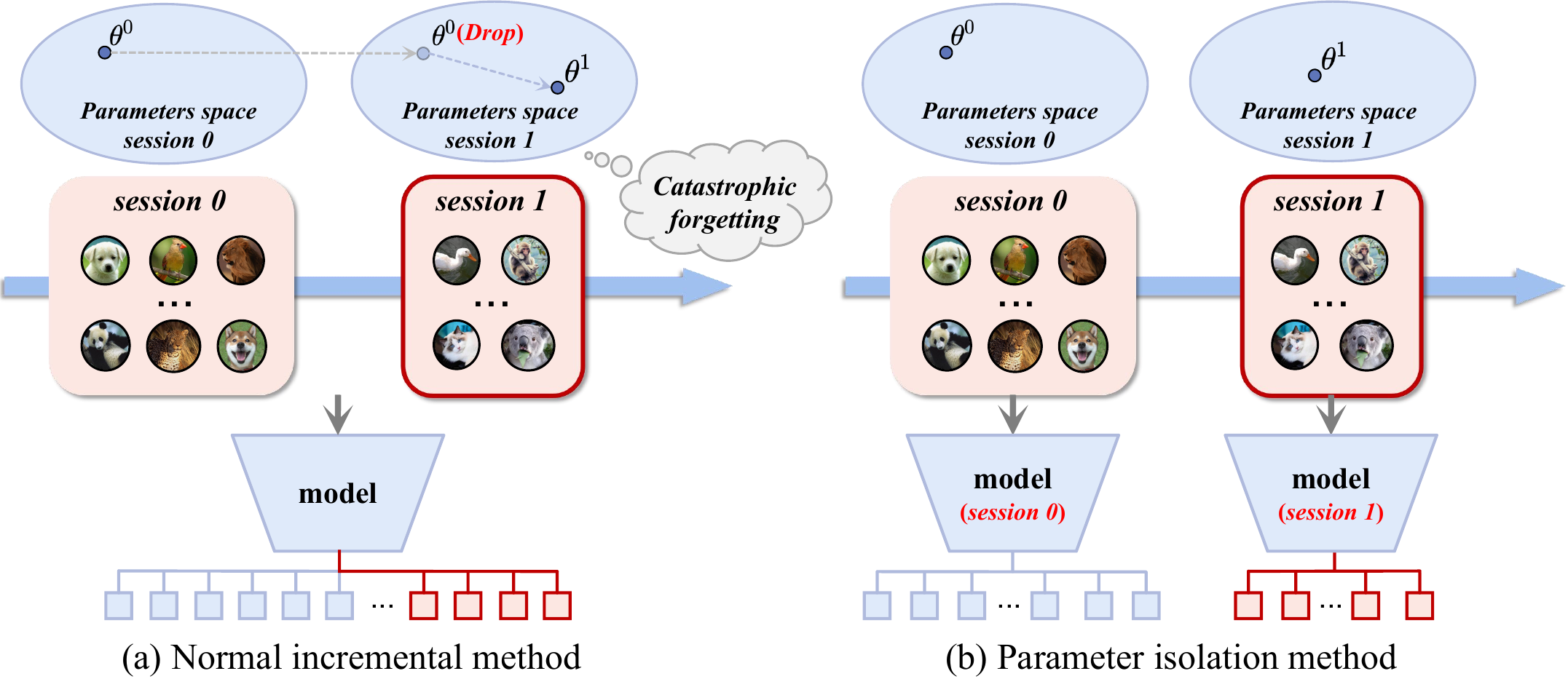}
    \caption{Comparing ordinary incremental learning method to parameter-isolation method. $\theta^t$ represents the model parameter on session $t$. (a) Model parameters changes trigger catastrophic forgetting (b) The model for each session learns the data independently and save the parameters separately.}
    \label{Introduction}
\end{figure}

Deep learning has achieved significant milestones in numerous large-scale computer vision tasks~\cite{he2016deep,radford2021learning,ramesh2022hierarchical}. These approaches generally invoke learning a mapping from samples to corresponding labels using extensive datasets. However, the mapping learned from specific data distribution is usually fixed and non-extended. \textit{i.e.}, a model can only recognize trained categories and cannot generalize to new categories. To enable the trained model to generalize effectively to new categories, \textit{Class Incremental Learning} (CIL) has received widespread attention~\cite{masana2022class,de2021continual}. CIL strives to enable models to continually learn different categories from data streams instead of a fixed dataset while preserving the capability to recognize previously encountered categories. Most CIL studies focus on letting models learn incrementally when the new class samples are sufficient. However, acquiring samples from new categories proves challenging and resource-intensive in many piratical scenarios.~\cite{zhou2022forward}. To tackle the challenge posed by the scarcity of samples from new categories, a task of more challenging and practical significance is denoted as FSCIL. In contrast to CIL, FSCIL must incrementally learn in situations with extremely limited labeled samples for new categories. The current FSCIL researches predominantly adhere to the traditional CIL paradigm, wherein a single model assimilates all data throughout the incremental process, sharing identical model parameters and decision boundaries. This learning paradigm presents significant challenges in conserving the model's old memory and deviates from the manner in which the human brain stores memories. As shown in Fig.~\ref{Introduction}, Acquiring knowledge of new categories inevitably alters parameters trained on old ones, causing catastrophic forgetting.

In contrast to this memory-burdened and forgettable learning paradigm, the human brain distributes learned knowledge in different areas of the cerebral cortex~\cite{kudithipudi2022biological,zhang2023few}. After exposure to a given task, a human can promptly associate it with the corresponding area of the cerebral cortex. This adaptive learning method is particularly well-suited for FSCIL because it can store memories separately. However, from the standpoint of biological and cognitive science, the challenges faced by deep neural networks in emulating the human cerebral cortex within FSCIL are principally manifested in the following two aspects. \textit{\textbf{i) How can the model imitate the cerebral cortex's subregional storage of acquired knowledge in memory?}} The unavailability of session information underscores the necessity for the model to possess the capability to recognize all previously learned categories. This limitation results in the necessity for the model to store the knowledge acquired across all sessions entirely within a unified memory area that cannot be segregated. \textit{\textbf{ii) How can one establish a mapping from task to memory?}} The session from which the test sample comes is unknown in FSCIL, establishing a mapping relationship between the samples and the model becomes unfeasible. \textit{i.e.}, The session to which the test sample belong i unknown. This implies that directly mapping the relevant areas of the cerebral cortex, as the human brain naturally does for a given task, is not feasible. This paper is dedicated to bridging the gap between the learning process of FSCIL and the way the human cerebral cortex stores memories by addressing the aforementioned challenges.

In this paper, to tackle the first challenge of enabling the classification system to store memories regionally, akin to the cerebral cortex, we adopt a parameter-isolation method, training distinct classification models for each session. In the conventional FSCIL, the knowledge pertaining to all sessions is consolidated within a singular model. This consolidation poses a risk of catastrophic forgetting. In our parameter-isolation method for FSCIL, each model exclusively stores knowledge acquired in its respective session, and the decision boundaries of these models operate independently, as shown in the Fig.~\ref{Introduction}(b). This approach to retaining previous knowledge effectively mirrors the memory partitioning storage mechanism observed in the human cerebral cortex. Furthermore, to mitigate the overfitting of new categories, we use the branch training strategy to train a series of models. To address the second challenge of establishing a mapping from samples to models, we predict the session information of the samples by UQ. Test samples are input into each model individually to obtain classification results along with corresponding uncertainty values. After, the appropriate classification result is selected based on the uncertainty value. UQ enables the model not only to produce its classification results but also to express the level of uncertainty associated with a given sample. The model gives low uncertainty to categories it recognizes and gives high uncertainty to categories it has never encountered. Relying on UQ, We successfully construct a memory map akin to the cerebral cortex. The model preserving the corresponding session memory is identified from a range of models for each test sample. 

To validate the effectiveness of proposed method, we conduct comprehensive comparative and ablation experiments on multiple benchmark datasets. The contribution of this work is summarized as follows:
 \begin{itemize}
     \item We propose a fresh viewpoint for FSCIL, emulating the memory storage mechanism of the human cerebral cortex. Our work is the first parameter-isolation method in FSCIL. Our approach achieves state-of-the-art performance on benchmark datasets.
     \item In the testing phase where session information is not accessible, we employ information entropy to quantify the uncertainty of a sample, enabling the inference of session information about the sample. 
     \item We employ a novel data augmentation strategy for real categories to generate virtual prototypes. This strategy significantly enhances the feature extraction capability of the backbone.
 \end{itemize}

 \section{Related Word}
\subsection{Few-shot Learning} 
FSL aims to recognize unseen objects from a few labeled samples~\cite{wang2020generalizing}. Currently, FSL researches can be mainly categorized as metric-based methods~\cite{koch2015siamese,vinyals2016matching,snell2017prototypical,yoon2019tapnet,li2019finding}, optimization-based methods~\cite{ravi2016optimization,finn2017model,sun2019meta,rusu2018meta}, and model-based methods~\cite{santoro2016meta,munkhdalai2017meta,munkhdalai2018rapid}. The metric-based approaches use the algorithm of nearest neighbor to learn the relationship between pairs of samples and assign labels to the samples with the smallest distance by calculating the distance between pairs of samples. \cite{munkhdalai2017meta} uses the mean of all samples in each category to represent the position of each category in the embedding space. In our approach, features are extracted using a pre-trained backbone and then classified by computing cosine similarity~\cite{vinyals2016matching} between the samples and the prototypes in the embedding space. 

\subsection{Class-incremental learning}
In the domain of class-incremental learning, the model consistently learns new knowledge from a data stream characterized by a diverse array of sample categories. This paper focuses on a class-incremental learning challenge that emerges when confronted with a constrained number of samples. iCaRL~\cite{rebuffi2017icarl} utilizes the nearest neighbor classifier to learn new categories and employs knowledge distillation as a strategy to preserve the knowledge of existing categories. LUCIR~\cite{hou2019learning} enhances performance in subsequent sessions by searching flat minima during the base training process. \cite{wu2022class} propose the same hypothesis to LUCIR: the base model plays a critical role in the overall performance of the incremental task. It establishes a robust feature extractor through training on base categories, subsequently freezing it. Then, for each incremental session, it fine-tunes the tail layers of the feature extractor and classifier individually. Inspired by~\cite{wu2022class}, in this paper, we employ the strategy of fine-tuning the tail of the feature extractor in FSCIL for incremental learning. Compared to fully freezing the feature extractor, this strategy allows the model to better adapt to new categories. 

\subsection{Few-shot class-incremental learning}
FSCIL represents a more challenging learning task compared to class-incremental learning. FSCIL aims to continuously acquire knowledge pertaining to new categories with a limited amount of labeled data while simultaneously preventing the inadvertent forgetting of previously learned categories~\cite{zhang2021few}. The problem setting of FSCIL is firstly proposed in TOPIC~\cite{tao2020few}. It utilizes the \emph{Neural Gas} (NG) network to stabilize the topological manifold features between new categories and old categories. CEC~\cite{zhang2021few} uses a graph neural network to update the classifier parameters in each incremental session based on the global knowledge of previous sessions. Simultaneously, it creates pseudo-incremental scenarios to optimize the graph neural network during base training. The pseudo-incremental approach has been used in many studies~\cite{zhu2021self,zhou2022forward}. Unlike previous approaches that only consider the performance of the current session, \cite{zhou2022forward} uses a method that enhances the scalability of the incremental model by compacting the embedding space through the inversion of virtual prototypes. In this paper, we crop and mix images with different categories to form images of virtual categories, relying on semantic information similar to that of real categories to generate virtual prototypes. 

\section{Problem Setting}
The fundamental aim of FSCIL is to perpetually learn novel categories in a few labeled samples, all the while safeguarding the retention of previously acquired knowledge. In FSCIL, the complete training dataset $D_{train}$ can be represented as $\{\mathcal{D}_{train}^{(0)}, \mathcal{D}_{train}^{(1)}, \mathcal{D}_{train}^{(2)}, \cdots, \mathcal{D}_{train}^{(N)} \}$, whever $\mathcal{D}_{train}^{(t)}$ represents the training data of $t$-th session, and $N$ is the total number of sessions. The corresponding label space is $\{ \mathcal{C}_{train}^{(0)}, \mathcal{C}_{train}^{(1)}, \mathcal{C}_{train}^{(2)}, \cdots, \mathcal{C}_{train}^{(N)} \}$. The samples and their labels from different sessions do not overlap, $\forall i, j \in \{ 0, 1, \cdots, N\}$ and $i \ne j$, $ \mathcal{D}_{train}^{(i)} \cap  \mathcal{D}_{train}^{(j)}= \varnothing$ and $ \mathcal{C}_{train}^{(i)} \cap  \mathcal{C}_{train}^{(j)}= \varnothing$. $\mathcal{D}_{train}^{(0)}$ is the base training dataset, which contains a large number of categories where each category has abundant training samples. $\mathcal{D}_{train}^{(i)} (i \in \{ 1, \cdots, N \}) $ is the incremental training dataset, which contains only a few samples, it is $N$ categories and each category contains $K$ samples, knows as $N$-way $K$-shot. During the training in the $t$-th session, the model only interacts with the samples and labels from the current session, $\mathcal{D}_{train}^{(i)}$ and $\mathcal{C}_{train}^{(i)}$. During the testing phase of the $i$-session, the evaluation involves all the categories that the model has already learned $\{ \mathcal{C}_{train}^{(0)} \cup \mathcal{C}_{train}^{(1)} \cup \cdots \cup \mathcal{C}_{train}^{(i)} \}$.

\section{Methodology}
\begin{figure}[t]
    \centering
    \includegraphics[width = 1.0\textwidth]{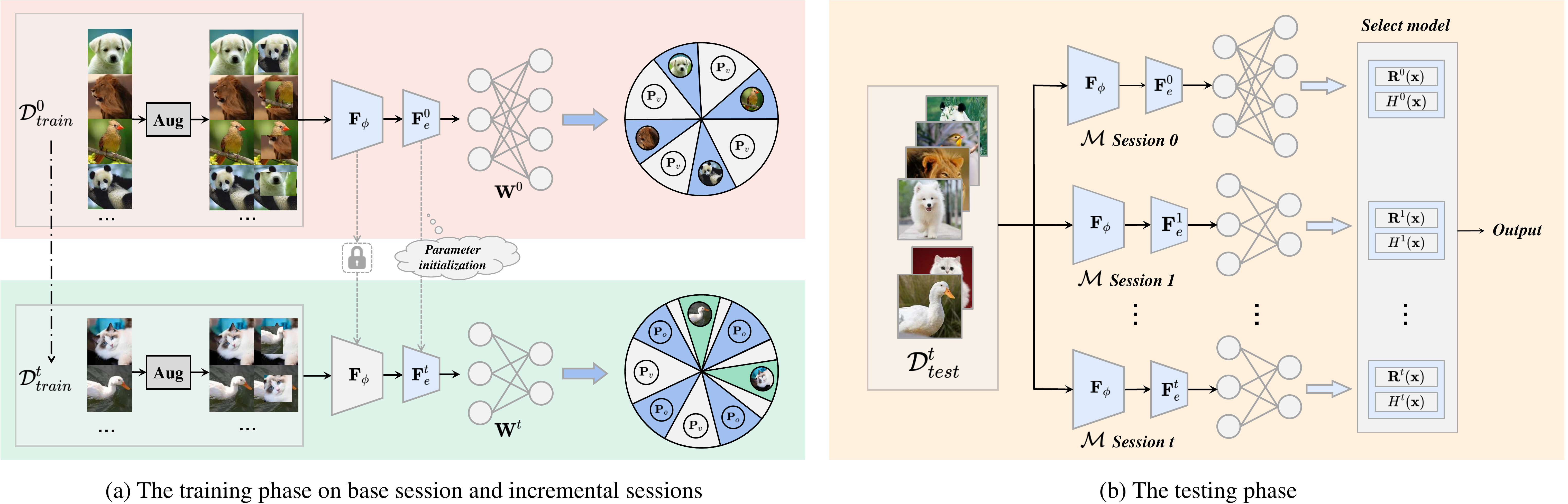}
    \caption{The overview of our proposed humankind memory-inspired FSCIL approach. (a) The CutMix data augmentation method enhances the feature extractor's performance by generating virtual prototypes. For session $t$, freeze the bulk $\mathbf{F}_\phi$ of the feature extractor, fine-tune the tail $\mathbf{F}_e^t$ and the classifier $\mathbf{W}^t$. (b) In the testing stage, the samples $\mathbf{x}$ are fed to a trained series of models outputting categories probability $\mathbf{R}^t(\mathbf{x})$ and uncertainty $H^t(\mathbf{x})$. The reliable result is then selected based on the value of $H^t(\mathbf{x})$. }
    \label{overflow}
\end{figure}

In this paper, We utilize a forward-compatible framework based on virtual prototype generation to enhance the backbone. It simulates incremental scenarios during the training phase of the base session to provide a prospective view of the model. In the incremental stage, we train the entire model using a representation learning decoupling strategy and a parameter isolation mechanism between sessions, saving parameters separately for each session. This is the first parameter isolation method in FSCIL. In the testing stage, we use UQ based on information entropy to select the best-matching classification result for the sample. The overall framework of our proposed method is shown in Fig.~\ref{overflow}.

\subsection{Forward compatible framework}
\begin{figure}[ht]
    \centering
    \includegraphics[width = 0.75\textwidth]{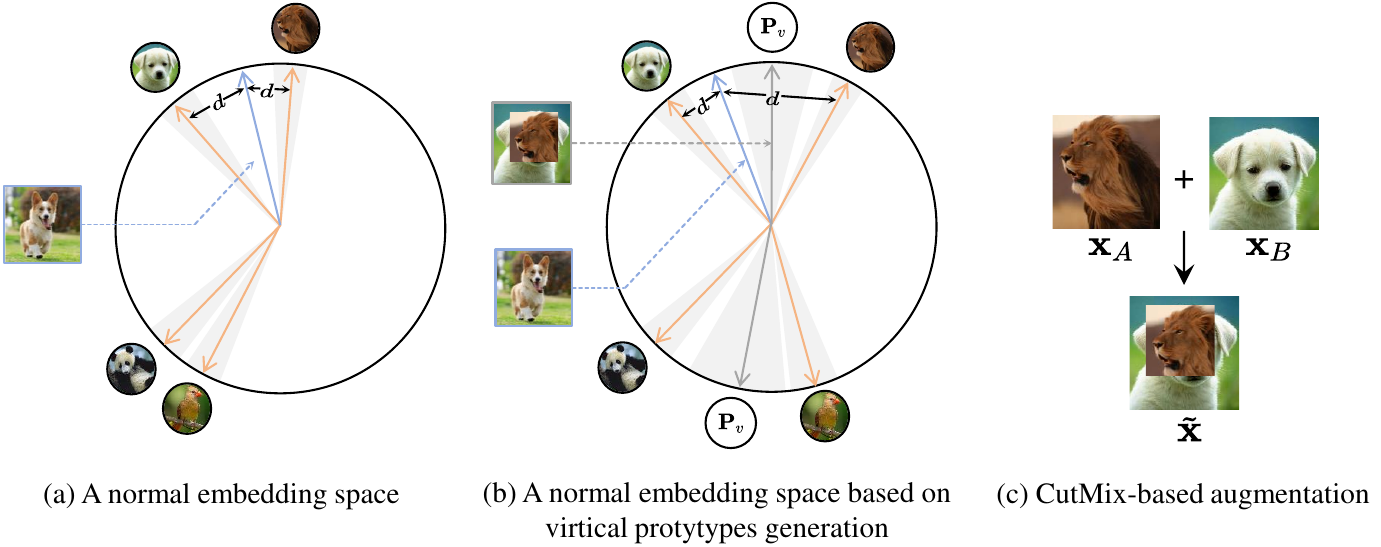}
    \caption{Virtual prototypes in embedding space. (a) Samples determine their category based on the closest prototype to them. Samples are more likely to be misclassified when the prototypes are too close to each other. (b) Inserting the virtual prototypes increases the separation between real prototypes and the embedding space's sparsity. (c) Procedure for virtual sample generation.}
    \label{embedding}
\end{figure}

FSCIL typically employs sufficient samples to optimize the empirical loss of the training data, thereby ensuring the robustness of the backbone~\cite{zhou2022forward}. Nevertheless, this approach can only learn decision boundaries for existing categories, making it incompatible between the new and old sessions. The incompatibility precludes the continuity between sessions, resulting in the insertion of new prototypes that disrupt the structure of the old session's embedding space. To facilitate the coordination of prototype distribution across the data stream as a whole rather than within the confines of individual sessions, we employ a forward-compatible framework based on virtual prototype generation. It utilizes virtual categories to simulate incremental scenarios during the training phase of the base session, allowing the embedding space to be reserved for future occurrences of the category, giving a forward-looking view of the entire model, and ensuring compatibility between different sessions. Our forward-compatible framework uses Cutmix~\cite{yun2019cutmix} to generate images of virtual categories. It takes an image that has been cropped and scaled and overlays it on an image from another category, as detailed in Fig.~\ref{embedding}. Eq.~\eqref{eqCutmix} shows the process of cropping and scaling image $\mathbf{x}_B$ onto image $\mathbf{x}_A$.
\begin{equation}\label{eqCutmix}
    \tilde{\mathbf x}=\mathbf{M} \odot \mathbf{x}_{A}+(\mathbf{1}-\mathbf{M}) \odot \mathbf{x}_{B}, \quad \mathbf{M}\in \{ 0, 1 \}^{H\times W}
\end{equation}

Where $\tilde{\mathbf{x}}$ denotes the image with the virtual category from the fusion of $\mathbf{x}_A$ and $\mathbf{x}_B$, $\mathbf{M}$ is a binary mask that drops image information, and $\odot$ represents pixel-by-pixel multiplication. We restrict $\mathbf{M}$ to half the original image size to integrate semantic features from both categories. The virtual samples generated by CutMix align closely with real categories in embedding space. It facilitates the recognition of samples by the distance metric, as the spacing between real prototypes is increased. Furthermore, the compression of existing prototypes allocates sufficient space for new categories. After training a sufficiently powerful feature extractor, samples from the real category are used to fine-tune the classifier to make the target category consistent with the real category.

\subsection{Decoupling in representation learning for parameters isolation}
\begin{figure}
    \centering
    \includegraphics[width=0.5\linewidth]{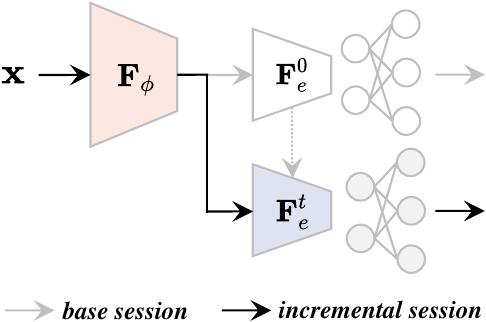}
    \caption{Decoupled training strategy in this paper. $\mathbf{x}$ represents the inputs. During incremental training, the majority of the feature extractor $\mathbf{F}_\phi$ is frozen to suppress catastrophic forgetting, while the tails $\mathbf{F}^t_e$ and classifiers undergo fine-tuning.}
    \label{branch}
\end{figure}

Previous FSCIL strategies decouple the feature extractor used to learn the representation and the classifier during the incremental process, freezing the parameters of the feature extractor to preserve the old class knowledge~\cite{zhang2021few, liu2022few}. However, the decoupling of representation learning and classifier obliterates the model's ability to learn representations of new categories. In fact, for a strong pre-trained feature extractor, fine-tuning its tails with new data can make it perform better on new sessions~\cite{wu2022class}. To facilitate the acquisition of more category-specific features during incremental sessions and the formation of more precise decision boundaries, we decouple the representation learning process, rather than merely decoupling the classifiers. Specifically, we partition the feature extractor into two parts, the main component $\mathbf{F}_\phi$ and the lesser layer at the tail $\mathbf{F}_e$. With sufficient data in the base session, we train $\mathbf{F}_\phi$, $\mathbf{F}_e^0$, and the classifier $\mathbf{W}^0$. Due to the scarcity of samples in the incremental process of FSCIL, fine-tuning the whole model can fall into an overfitting dilemma. Therefore, in incremental sessions, the parameters of $\mathbf{F}_\phi^t$ are frozen to preserve the fundamental representational capabilities, and the $\mathbf{F}_e^t$ and $\mathbf{W}^t$ $(t \geq 1)$ undergo fine-tuning to learn the representation of new categories. Fig.~\ref{branch} illustrates this decoupled training strategy. Our decoupling strategy for representation learning permits the model to learn features about new categories during each incremental session, rather than extracting feature vectors using parameters on the base session. To allow decoupled models to avoid catastrophic forgetting while learning new categories, we use a parameter isolation strategy to keep separate model parameters for each session. The parameters of $\mathbf{F}_\phi$ are shared for each session, and $\mathbf{F}_e^t$ and $\mathbf{W}^t$ are stored separately for each session.

The feature vector $\mathbf{p}_i$ extracted from the image $\mathbf{x}_i$ by the feature extractor is shown in Eq.~\eqref{eqP}. The feature extractor extracts all training samples of class $k$ to generate feature vectors and computes the mean to obtain the feature prototype $\overline{\mathbf{p}^k}$ for that class. The feature prototypes are calculated as shown in Eq.~\eqref{eqPrototype}. In base session and incremental sessions, the cross-entropy loss function as shown in Eq.~\eqref{eqCE} is used for training.

\begin{equation}\label{eqP}
    \mathbf{p}_i = \mathbf{F}_e(\mathbf{F}_\phi(\mathbf{x}_i)) 
\end{equation}

\begin{equation}\label{eqPrototype}
    \overline{\mathbf{p}^k} = \frac{1}{|\mathcal{C}(k)|}\sum_{i=1}^{|\mathcal{C}(k)|}\mathbf{p}_i^k
\end{equation}

\begin{equation}\label{eqCE}
    \mathcal{L}_{CE} = \sum_{j=1}^N\log(\frac{e^{\cos<\overline{\mathbf{p}^k}, \mathbf{p}_j^k>}}{\sum_{i=1}^{|\mathcal{C}|}e^{\cos<\overline{\mathbf{p}^i}, \mathbf{p}_j^k>} })
\end{equation}

\subsection{Model selection based on uncertainty qualification}
During FSCIL testing, without access to session labels, mapping samples to the correct model poses a critical challenge. During testing stage, we have a feature extractor backbone $\mathbf{F}_\phi$, a sequence of feature extractor tails $\{\mathbf{F}_e^0, \mathbf{F}_e^1, \dots, \mathbf{F}_e^n\}$ and a sequence of classifier $\{ \mathbf{W}^0, \mathbf{W}^1, \dots, \mathbf{W}^n \}$. Choosing the right one from a range of models for a test sample is critical. UQ can measure the model's certainty level with fed samples, producing uncertainty values for classification. In this study, we use information entropy to measure the uncertainty of the fed samples. The information entropy is calculated as shown in Eq.~\eqref{EqIE}.
\begin{equation}\label{EqIE}
    H^t(\mathbf{x}) = -\sum_{c=1}^{|\mathcal{C}^t|}p(o^c)\log p(o^c)
\end{equation}

Where $H^t(\mathbf{x})$ denotes the information entropy of the model in session $t$ for the fed sample $\mathbf{x}$, $|\mathcal{C}^t|$ denotes the total number of categories in session $t$, $p(o^c)$ represents the probability of class $c$ for fed sample $\mathbf{x}$. During the testing stage, the samples are fed into all session models. Simultaneously, each model computes the uncertainty of the fed samples based on Eq.~\eqref{EqIE}. $p(o^c)$ is obtained by mapping the neural network output to a probability. Finally, the model with the minimum uncertainty is chosen as the classification model for the given sample, and the predicted class from that model is accepted.

\subsubsection{Categories imbalance in uncertainty qualification}
Identical target categories for both classification systems are crucial to render the results of the UQ meaningful and comparable for analysis. However, in FSCIL's setting, base sessions comprise many categories, while incremental sessions feature fewer. To maintain consistency in target categories between the base session model and incremental session models during UQ, we partition the probabilities generated by the base session model to $N_{sub}$ sub-results, $N_{sub}$ as shown in Eq.~\ref{eqSub}.
\begin{equation}\label{eqSub}
    N_{sub} = \frac{|\mathcal{C}^0|}{|\mathcal{C}^i|}, \quad |\mathcal{C}^i| = |\mathcal{C}^j| (1 \le i, j \le n)
\end{equation}

$|\mathcal{C}^0|$ and $|\mathcal{C}^i|$ represent the number of categories in the base session and the incremental session, respectively. The output of the base session model has a total of $N_{sub}$ sub-results, where each sub-results contains the probability of $|\mathcal{C}^i|$ categories. We quantify the uncertainty for each sub-result and select the smallest as the UQ result for the whole base session model. The uncertainty values obtained before the sub-result division are shown in Eq.~\eqref{eqU}. The uncertainty values obtained after dividing the results of the base session classification are shown in Eq.~\eqref{eqU2}.
\begin{equation}\label{eqU}
    \mathbf{U} = \{ H^0(\mathbf{x}_i), H^1(\mathbf{x}_i), \cdots, H^n(\mathbf{x}_i) \}
\end{equation}
\begin{equation}\label{eqU2}
    \mathbf{U} = \{ H^0_{1}(\mathbf{x}_i), \cdots, H^0_{N_{sub}}(\mathbf{x}_i), H^1(\mathbf{x}_i), \cdots, H^n(\mathbf{x}_i) \}
\end{equation}

Where $ \mathbf{U}_{\mathrm{base}} = \{ H^0_{1}(\mathbf{x}_i), \cdots, H^0_{N_{sub}}(\mathbf{x}_i) \}$ denotes the set of all sub-results from the base session model. The minimum value in $ \mathbf{U}_{\mathrm{base}}$ as the uncertainty value of the base session model participates in the UQ containing all session models. This sub-results partitioning strategy for the base session model effectively tackles the category imbalance between the base and incremental categories. This is achieved by aligning the categories in the sub-results of the base session with the incremental class.

\section{Experiments}

\subsection{Dataset}

CIFAR-100 contains 100 classes with 600 $32\times 32$ RGB images per class, of which 500 are used for training and 100 for testing, and the entire dataset has 600,000 images. 60 classes are used as base classes, 40 as incremental classes, and eight sessions are in the incremental stage. The data in each incremental session appears as the 5-way 5-shot. \textit{mini}-ImageNet is a subset of ImageNet that contains 100 classes, each containing 600 $84\times 84$ RGB images. 60 classes are used as base classes, and 40 are incremental classes. The 40 classes are evenly distributed into eight sessions, with five classes per session. The data in each incremental session appears as the 5-way 5-shot.

\subsection{Implementation Details}
Same as other FSCIL studies~\cite{zhang2021few,liu2022few,zhuang2023gkeal}, we employ ResNet-18 as the backbone to validate the performance of our proposed method on the benchmark datasets. We follow the setting of~\cite{zhuang2023gkeal,zhang2021few} for CIFAR-100 and \textit{mini-}ImageNet, and we use randomly initialized model parameters. The entire feature extraction section of ResNet-18 is partitioned into four blocks, each consisting of three convolution layers. The first three blocks are $\mathbf{F}_\phi$, where the parameters are frozen after training in the base session. The last block in the tail is $\mathbf{F}_e$, and it continuously updates the parameters during incremental sessions. 

\subsection{Comparison with the state-of-the-art methods}
\begin{figure}[ht]
    \centering
    \includegraphics[width = 0.8\textwidth]{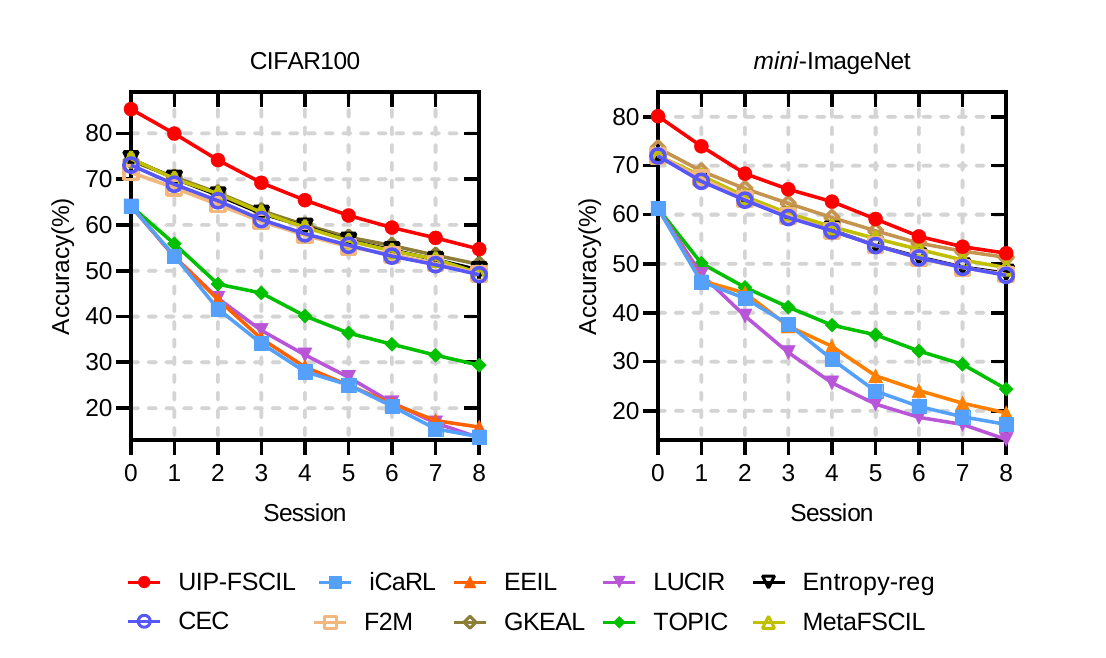}
    \caption{Average accuracy in each session of two benchmark datasets: CIFAR-100, \textit{mini-}ImageNet.}
    \label{performance}
\end{figure}

To validate the state-of-the-art performance of our method, we conduct comparative experiments with contemporary approaches on two benchmark datasets: CIFAR-100 and \textit{mini}-ImageNet. The comparative performance of our method compared to other FSCIL methods is shown in Fig.~\ref{performance}. Furthermore, the detailed performance for CIFAR-100 is shown in Tab.~\ref{acccifar100}.

\begin{table}[ht]
    \renewcommand{\arraystretch}{1}
    \centering
    \caption{The accuracy of each session among the compared methods on CIFAR-100. \textbf{AA} is average accuracy.}
    \label{acccifar100}
     \resizebox{\linewidth}{!}{
    \begin{tabular}{lccccccccccc}
        \toprule[1.5pt]
        \multirow{2}{*}{Method} & \multirow{2}{*}{Venue}  & \multicolumn{9}{c}{Accuracy in each session (\%)} & \multirow{2}{*}{AA} \\
        \cline{3-11}
        ~ & ~ & 0 & 1 & 2 & 3 & 4 & 5 & 6 & 7 & 8 \\
        \midrule
        iCaRL~\cite{rebuffi2017icarl} & CVPR17 & 64.10 & 53.28 & 41.69 & 34.13 & 27.93 & 25.06 & 20.41 & 15.48 & 13.73 & 32.86\\
        EEIL~\cite{castro2018end} & ECCV18 & 64.10 & 53.11 & 43.71 & 35.15 & 28.96& 24.98 & 21.01 & 17.26 & 15.85 & 33.79 \\
        LUCIR~\cite{hou2019learning} & CVPR19 & 64.10 & 53.05 & 43.96 & 36.97 & 31.61 & 26.73 & 21.23 & 16.78 & 13.54 & 34.21\\
        TOPIC~\cite{tao2020few} & CVPR20 & 64.10 & 55.88 & 47.07 & 45.16 & 40.11 & 36.38 & 33.96 & 31.55 & 29.37 & 42.62 \\
        CEC~\cite{zhang2021few} & CVPR21 & 73.07 & 68.88 & 65.26 & 61.19 & 58.09 & 55.57 & 53.22 & 51.34 & 49.14 & 59.52 \\
        F2M~\cite{shi2021overcoming} & NIPS21 & 71.45 & 68.10 & 64.43 & 60.80 & 57.76 & 55.26 & 53.53 & 51.57 & 49.35 & 59.13\\
        MetaFSCIL~\cite{chi2022metafscil} & CVPR22 & 74.50 & 70.10 & 66.84 & 62.77 & 59.48 & 56.52 & 54.36 & 52.56 & 49.47 & 60.73\\
        Entropy-reg~\cite{liu2022few} & ECCV22 & 74.40 & 70.20 & 66.54 & 62.51 & 59.71 & 56.58 & 54.52 & 52.39 & 50.14 & 60.77\\
        GKEAL~\cite{zhuang2023gkeal} & CVPR23 & 74.01 & 70.47 & 67.01 & 63.08 & 60.01 & 57.30 & 55.50 & 53.39 & 51.40 & 61.35\\
        \hline
        Ours & - & \textbf{85.30} & \textbf{79.94} & \textbf{74.16} & \textbf{69.19} & \textbf{65.40} & \textbf{62.09} & \textbf{59.40} & \textbf{57.16} & \textbf{54.73} & \textbf{67.44} \\ 
        \bottomrule[1.5pt]
    \end{tabular}}
\end{table}

Based on the experimental results, our approach attains satisfactory performance for both CIFAR-100 and \textit{mini}-ImageNet, demonstrating excellent in base session and incremental sessions. In CIFAR-100, our method outperforms GKEAL~\cite{zhuang2023gkeal} by 3.73\% in the final session, with an average accuracy improvement of 6.09\%. The experimental results unequivocally demonstrate that our method outperforms other state-of-the-art methods for FSCIL datasets. Our proposed parameter-isolation approach in FSCIL effectively mitigates the issue of catastrophic forgetting. Simultaneously, the experimental results substantiate that the UQ technique can construct a mapping from samples to models. This learning method offers a fresh perspective for the FSCIL field.

\subsection{Ablation Study}
To assess the efficacy of our methods, we conduct ablation experiments on CIFAR-100. The detailed results are provided in Tab.~\ref{ablation}. Post application of the forward-compatible framework, accuracy on base session rose to 82.93\%. This suggests that the method effectively enhances backbone feature extraction. However, deploying parameter isolation with model selection yielded unsatisfactory performance due to category imbalance in UQ. We address this with the SR strategy during testing, resulting in AA increasing from 58.38\% to 65.27\%. This underscores the importance of addressing category imbalance in UQ within the FSCIL domain. Further, fine-tuning the model with real categories after forward compatibility increased AA by 2.17\%. In summary, our method raised the last session accuracy from 49.04\% to 54.73\% and AA from 62.88\% to 67.44\%. This indicates that our method effectively resists forgetting and generalizes to new categories.

\begin{table}[ht]
    \renewcommand{\arraystretch}{1}
    \centering
    \caption{The comprehensive ablation experiment on CIFAR-100. \textbf{FC} is the forward forward-compatible framework. \textbf{DR} is decoupling in representation learning. \textbf{MS} is the model selection based on UQ \textbf{SR} is the sub-results of the classification model in the base session. \textbf{FT} is the fine-tuning for the category's number of classifiers.}
    \label{ablation}
     \resizebox{\linewidth}{!}{
    \begin{tabular}{cccccccccccccccc}
        \toprule[1.5pt]
        \multirow{2}{*}{\textbf{FC}} & \multirow{2}{*}{\textbf{DR}} & \multirow{2}{*}{\textbf{MS}} & \multirow{2}{*}{\textbf{SR}} & \multirow{2}{*}{\textbf{FT}} & \multicolumn{9}{c}{Accuracy in each session (\%)} & \multirow{2}{*}{AA} \\
        \cline{6-14}
        ~ & ~ & ~ & ~ & ~ & 0 & 1 & 2 & 3 & 4 & 5 & 6 & 7 & 8 \\
        \midrule
        ~ & ~ & ~ & ~ & ~ & 80.92 & 75.86 & 69.81 & 65.45 & 61.13 & 57.39 & 54.41 & 51.99 & 49.04 & 62.88\\
        \checkmark & ~ & ~ & ~ & ~ & 82.93 & 76.35 & 71.22 & 66.32 & 61.88 & 58.65 & 55.71 & 52.71 & 49.37 & 63.90\\
        \checkmark & \checkmark & \checkmark & ~ & ~ & 82.93 & 72.06 & 64.45 & 59.03 & 55.51 & 51.49 & 49.08 & 46.95 & 43.98 & 58.38 \\
        \checkmark & \checkmark & \checkmark & \checkmark & ~ & 82.93 & 77.60 & 71.90 & 67.02 & 63.62 & 59.92 & 56.94 & 54.73 & 51.79 & 65.27 \\
        \checkmark & \checkmark & \checkmark & \checkmark & \checkmark & 85.30 & 79.94 & 74.16 & 69.19 & 65.40 & 62.09 & 59.40 & 57.16 & 54.73 & 67.44 \\
        \bottomrule[1.5pt]
    \end{tabular}}
\end{table}

\section{Conclusion}
This paper proposes a FSCIL method that emulates memory storage in the human cerebral cortex. This is the first parameter isolation method in FSCIL. We train distinct classification models for each session to uphold the performance of the entire classification system on old categories. During the testing, we quantify the uncertainty of the samples in deriving session labels, facilitating the selection of the appropriate classification model. By emulating the memory storage mechanism of the cerebral cortex, our approach renders the learning process of FSCIL more aligned with the human learning style, offering a novel perspective to the field of FSCIL. In the future, we aim to develop UQ methods that are better suited for limited samples and meticulously consider detailed feature differences.

\section{Acknowledgements}
This work was supported by the Outstanding Youth Project from the Hunan Provincial Department of Education under Grant 24B0865. and the China Scholarship Council under Grant 202206110005. Renye Zhang and Yimin Yin are with the National University of Defense Technology and Hunan First Normal University, Changsha, China (e-mail: renyezhang2016@163.com, yinyimin1@126.com). Jinghua Zhang is with the National University of Defense Technology, Changsha, China (e-mail: zhangjingh@foxmail.com). Yimin Yin is the co-first author. Jinghua Zhang is the corresponding author.

\bibliographystyle{elsarticle-num}
\bibliography{refers}
\end{document}